\documentclass{article}
\usepackage{ijcai18}

\usepackage{times}
\usepackage{xcolor}
\usepackage{soul}
\usepackage[utf8]{inputenc}
\usepackage[small]{caption}

\usepackage{amsmath}
\usepackage{amsthm}
\usepackage{graphicx}

\usepackage{amssymb}

\title{Constrained shortest path search with graph convolutional neural networks}

\author{
Kevin Osanlou$^{1,3}$, 
Christophe Guettier$^1$, 
Andrei Bursuc$^2$,
Tristan Cazenave$^3$, 
Eric Jacopin$^4$, 
\\ 
$^1$ Safran Electronics \& Defense\\
$^2$ Safran SA\\
$^3$ LAMSADE, Université Paris-Dauphine\\
$^4$ MACCLIA, Ecoles de Saint-Cyr Coëtquidan\\
kevin.osanlou@safrangroup.com
christophe.guettier@safrangroup.com
andrei.bursuc@safrangroup.com\\
tristan.cazenave@lamsade.dauphine.fr
eric.jacopin@st-cyr.terre-net.defense.gouv.fr
}

\begin{document}

\maketitle

\def\etal{\emph{et al.}}
\def\ie{\emph{i.e.}}
\def\eg{\emph{e.g.}}

\newcommand{\real}{\mathbb{R}}
\newcommand{\tran}{^{\mathrm{T}}}

\def\ab{\mathbf{a}}
\def\xb{\mathbf{x}}
\def\yb{\mathbf{y}}
\newcommand{\thetab}{{\boldsymbol \theta}}

\begin{abstract}

Planning for Autonomous Unmanned Ground Vehicles (AUGV) is still a challenge, especially in difficult, off-road critical situations. Automatic planning can be used to reach mission objectives, to perform navigation or maneuvers. Most of the time, the problem consists in finding a path from a source to a destination, while satisfying some operational constraints. In a graph without negative cycles, the computation of the single-pair shortest path from a start node to an end node is solved in polynomial time. Additional constraints on the solution path can however make the problem harder to solve. That becomes the case if we require the path to pass by a few  mandatory nodes without requiring a specific order of visit. The complexity grows exponentially as we require more nodes to visit. In this paper, we focus on shortest path search with mandatory nodes on a given connected graph. We propose a hybrid model that combines a constraint-based solver and a graph convolutional neural network to improve search performance. Promising results are obtained on realistic scenarios.

\end{abstract}

\section{Introduction}
\label{sec:intro}

Autonomous unmanned ground vehicle (AUGV) operations are constrained by terrain structure, observation abilities, 
embedded resources. Missions must be executed in minimal time, while meeting objectives. 
This is the case, for examples, in disaster relief, logistics or area surveillance, where maneuvers must consider terrain knowledge.
In most of applications, the AUGV ability to maneuver in its environment has a direct impact on operational efficiency.  

AUGV integrates several perception capabilities (on-line mapping, geolocation, optronics, LIDAR) in order to update its situation awareness on-line. 
This knowledge is used by various on-board planning layers to maintain mission goals, provide navigation waypoints and dynamically construct platform maneuvers. 
Resulting actions and navigation plans are used for controlling the robotic platform. Once the plans are computed, the AUGV automatically manages its trajectory and follows the navigation waypoints using control algorithms and time sequence.
Such autonomous system architectures are challenging, especially because some mission data (terrain, objectives, available resources) are known off-line and others are acquired on-line.
The planning problem also involves technical actions (observations, measurements, communications, \dots) to realize on some mandatory waypoints along the navigation plan \cite{guettier_lucas_2016}.

\begin{figure}
\centering
\includegraphics[scale=0.7]{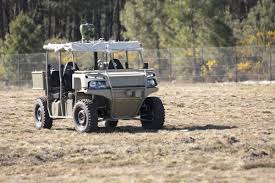}
\caption{The {\it eRider}, developed by SAFRAN, is an all-purpose optionally piloted vehicle that can patrol a given area, observe at long range or carry goods for various off-road applications in difficult environments. These specific vehicles can be either piloted or turned instantaneously into AUGV.}
\label{fig_ugv}
\end{figure}

Figure~\ref{fig_ugv} presents the {\it eRider}, a UGV that has high mobility abilities, and is able to perform cross-over maneuvers on difficult terrain for exploring disaster areas.

The paper focuses on the automatic planning algorithms, and more specifically, on the ability to guide the problem solving with machine learning techniques.

For such problems, classical robotic systems integrate A* algorithms \cite{Har68}, as a best-first search approach in the space of available paths.
For a complete overview of static algorithms (such as A*), replanning algorithms (D*), anytime algorithms (e.g. ARA*), and anytime replanning
algorithms (AD*), see \cite{Ferguson-2005-9201}. Representative applications to autonomous systems can be very realistic, as reported in \cite{Meu09} and \cite{Meu11}.

Algorithms stemming from A* can handle some heuristic metrics but can become complex to develop when dealing with several constraints simultaneously like mandatory waypoints and distance metrics. 
Our approach combines a Constraint Programming (CP) method, with novel machine learning techniques. CP provides a powerful baseline to model and solve combinatorial and / or constraint satisfaction problems (CSP).
It has been introduced in the late 70s \cite{Lauriere1978} and has been developed until now \cite{Hen98,AW03,Carlsson2015}, with several real-world autonomous system applications, 
in space \cite{BGP00,SAHL15}, aeronautics \cite{GAVP02,guettier_lucas_2016} and defense \cite{Guettier_and_all_2015}. 

Convolutional neural networks have proven to be very efficient when it comes to image recognition. 
They use a variation of multilayer perceptrons designed to require minimal preprocessing, and are capable of detecting complex patterns in the image. 
In this paper, we are dealing with graphs representing maneuvers or off-road navigation. 
Instead of convolutional neural networks, the paper focuses on graph convolutional neural networks, a variant aimed at learning complex patterns in graphs, and capable of making decisions to solve path-related problems. These machine learning algorithms are combined  with constraint solving methods for planning in the article.  

The paper is organized as follows. The first section (\S~\ref{sec:context}) describes AUGV mission, environment and applications.
The following section (\S~\ref{sec:problem_resolution}) presents the CP approach, problem formalization and resolution.
Section (\S~\ref{sec:learning}) describes the graph-based learning algorithm and section (\S~\ref{sec:data_generation}) the data generation schema for training.
Last sections (\S~\ref{sec:results}) and (\S~\ref{sec:conclusion}) discuss results, highlight further work and conclude.
Related works are provided along the different sections.

\section{Context and problem presentation}
\label{sec:context}

In modern AUGV architectures, path planning is performed on the fly, responding to an operator demand (mission update), a terrain obstacle, or free space discovery. 
In logistics, such operator demands correspond to movement requests for pick-up and delivery, while in area surveillance or disaster relief, 
the platform must reccon some specific areas.

\begin{figure}
\centering
\includegraphics[scale=0.5]{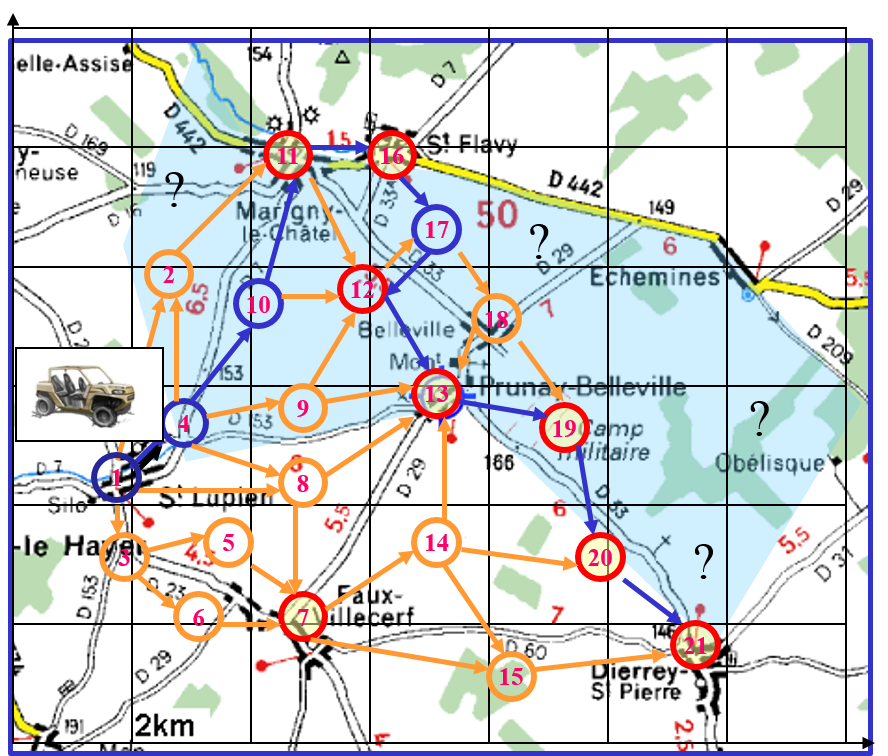}
\caption{Search and rescue mission.  Potential paths for a manned vehicle and a AUGV throughout a flooded area between the Seine and the Vanne rivers in the area of Troyes, France. The blue area represents the expected flooded area. Red circles represent possible waypoints while blue circles  represent mandatory ones. Orange edges represent trafficability and the blue ones a potential optimal solution.}
\label{fig_nav}
\end{figure}

Figure~{\ref{fig_nav}} shows a flooded area and possible paths to assess disaster damages. Possible paths are defined using a graph representation, where edges and nodes represent respectively ground mobility and accessible waypoints.
With the given applications, graphs are defined during mission preparation, by terrain analysis and situation assessment. The UGV system responds to an operator demand by finding a route from a starting point to a destination, and by maneuvering through mandatory waypoints.
During the mission, some paths may not be trafficable or new flooded areas to investigate may be identified.

These constraints depend on both environment situation and mission objectives:

\begin{itemize}
\item{Environment situation:} new muddy areas can occur, affecting cross-over duration between two waypoints or increasing the risk of losing platform control.  

\item{Mission objectives:} mandatory waypoints are imposed by the user and can correspond, for instance, to observation spots.
\end{itemize}

In figure~\ref{fig_nav}, the UGV starts from its initial position (blue circle) and must visit waypoints $\{W_1,W_2,W_3,W_4\}$. 
On the disaster relief scenario, areas in blue are flooded and the disaster perimeter must be evaluated by the vehicle. All nodes circled in red have to be visited, where refugees and casualties are likely to be found. The main criterion to minimize is the
global traverse duration that meets all visit objectives. A typical damage assessment would require up to 10 mandatory nodes to visit. Using the UGV system, the remote operator can update the situation and set objectives accordingly, which usually yields replanning events. To achieve high operational efficiency, path planning must be solved 'just in time' to execute maneuvers smoothly. 
However, such planning problems rapidly become NP-hard and state-of-the-art solvers may involve heavy processing loads, whereas a solution is required right away. 

Our proposed approach is to learn problem invariants from the terrain structure in order to accelerate a model-based planner.
This can be done by training a neural network on several problem instances. Using feedback from the neural network, the model-based planner can efficiently solve new problem instances. Learning from terrain data can be performed off-line, such that the knowledge acquired can be used on-line to accelerate the on-board planner.  However, critical missions such as disaster relief necessitate to deploy the vehicle rapidly and it is not possible to process heavy datasets of problem instances.

Two scenarios are considered in this paper, leading to two evaluation benchmarks. The first scenario, $b_1$, is a fine grain maneuver on a muddy terrain of small size (referred as benchmark {\it 'maneuver'}), while the second one, $b_2$, is a coarse mapping (referred as benchmark {\it 'exploration'}) of a disaster risk on a shore environment nearby a city.

\section{Problem formalization and resolution approach}
\label{sec:problem_resolution}

This section presents the global constraint programming approach to solve complex planning problems.
Let $\mathcal{G}=(\mathcal{V},\mathcal{E})$ be a connected graph. A typical instance $I$ of the kind of path-planning problem we consider is defined as follows:  
 $$I=(s,d,M)$$ where:
\begin{itemize}
\item $s \in V$ is the start node in graph $\mathcal{G}$,
\item $d \in V$ is the destination node in graph $\mathcal{G}$,
\item $M \subset V$ is a set of distinct mandatory nodes that need to be visited at least once, regardless of the order of visit.
\end{itemize}

In order to solve instance $I$, one has to find a shortest path from node $s$ to node $d$ that passes by each node in $M$ at least once. 
There is no limit to how many times a node can be visited in a path. Since the graph is connected, there is a solution path to every existing instances.
Let $A$ be the adjacency matrix of graph $\mathcal{G}$, used in (\S \ref{sec:learning}) by our neural network, defined as follows:

\[
    A_{vv'}= 
\begin{cases}
    \ 0, & \text{if there is no edge from node $v$ to node $v'$} \\
    \ w_{vv'},   & \text{otherwise, and the weight of the edge is $w_{vv'}$}
\end{cases}
\]

Let $A'$ be the cost matrix of graph $\mathcal{G}$, used by our path-planning solver, defined as follows:

\[
    A'_{vv'}= 
\begin{cases}
    \ \infty, & \text{if there is no edge from node $v$ to node $v'$} \\
    w_{vv'},   & \text{otherwise, and the weight of the edge is $w_{vv'}$}
\end{cases}
\]

\subsection{Constraint Programming for Navigation and Maneuver Planning}

In our approach, planning is achieved using Constraint Programming (CP) techniques, under a model-based development approach.
In CP, it is possible to design global search algorithms that guarantee completeness and optimality.
A CSP, formulated within a CP environment, is composed of a set of variables, their domains and algebraic constraints, that are based on problem discretization.
With CP, a declarative formulation of the constraints to satisfy is provided which is decoupled from the search algorithms, so that both can be worked out independently.
The CSP formulation and search algorithms exposed in the paper are implemented with the CLP(FD) domain of SICStus Prolog library \cite{Carlsson2015}.
It uses the state-of-the-art in discrete constrained optimization techniques Arc Consistency-5 (AC-5) \cite{DH91,HDT92} for constraint propagation, managed by CLP(FD) predicates.
With AC-5, variable domains get reduced until a fixed point is reached by constraint propagation. 
The search technique is hybridized by statically defining the search exploration structure using probing and learning on multiple problem instances. 
This approach, named probe learning, relies on several instances of a problem to build up the search tree structure.

\subsection{Planning model with Flow Constraints}

The set of flow variables $\varphi_u \in \{0,1\}$
models a possible path from $start \in X$ to $end \in X$, where an edge $u$
belongs to the navigation plan if and only if a decision variable
$\varphi_u = 1$, $0$ otherwise. The resulting navigation plan, can be represented as $\Phi=\{u|~u\in
U,~\varphi_u=1\}$. From an initial position to a requested final one, path consistency is enforced
by flow conservation equations, where $\omega^+(x) \subset U$ and
$\omega^-(x) \subset U$ represent respectively outgoing and incoming edges from vertex
$x$. Since flow variables are $\{0,1\}$, equation (\ref{cons_path}) ensures path connectivity and uniqueness while equation (\ref{start_end_path})
imposes limit conditions for starting and ending the path: 

\begin{small}
\begin{eqnarray}
\underset{u~\in~\omega^+(x)}\sum \varphi_u = \underset{u~\in~\omega^-(x)}\sum \varphi_u \leq N\label{cons_path}\\
\underset{u~\in~\omega^+(start)}\sum \varphi_u~=~1,~~\underset{u~\in~\omega^-(end)}\sum \varphi_u~=~1,\label{start_end_path}
\end{eqnarray}
\end{small}

These constraints provide a linear chain alternating pass-by waypoint and navigation along the graph edges.
Constant $N$ indicates the maximum number of times the vehicle can pass by a waypoint.
With this formulation, the plan may contain cycles over several waypoints.
Mandatory waypoints are imposed using constraint (\ref{mandatory_path}). 
The total path length is given by the metric (\ref{cumulative}), and
we will consider the path length as the optimization criterion to minimize in the context of this paper:

\begin{small}
\begin{eqnarray}
\forall i \in M \underset{u~\in~\omega^+(i)}\sum \varphi_u~\geq~1\label{mandatory_path}\\
D_v=\underset{v'v~\in~\omega^-(v)}\sum \varphi_{v'v} w_{vv'} \label{cumulative}
\end{eqnarray}
\end{small}

\subsection{Global search algorithm}

The global search technique under consideration guarantees completeness, solution optimality and proof of
optimality. It relies on three main algorithmic components:

\begin{itemize}
\item Variable filtering with correct values, using specific labeling predicates to instantiate problem domain variables.
AC being incomplete, value filtering guarantees search completeness.
\item Tree search with standard backtracking when variable instantiation fails.
\item Branch and Bound (B\&B) for cost optimization, using minimize predicate.
\end{itemize}

Designing a good search technique consists in finding the right variables ordering and value filtering, accelerated by domain or generic heuristics. 
A static probe provides an initial variable selection ordering, computed before running the global branch and bound search.
Note that in general probing techniques \cite{Els00}, the order can be redefined within the search structure \cite{Rum01}.
Similarly, in our approach, the variable selection order provided by the probe can still be iteratively updated by the labeling strategy that makes use of other variable selection heuristics.
Mainly, first fail variable selection is used in addition to the initial probing order. 
These algorithmic designs have already been reported with different probing heuristics \cite{guettier_lucas_2016}, such as A* or meta-heuristics such as Ant Colony Optimization \cite{Luc10a},\cite{LGS09}.
In our design, the search is still complete, guarantying proof of optimality, but demonstrates efficient pruning.
Instead of these heuristics techniques, we choose to train with multiple instances the probing mechanism that provides a tentative variable order to the global search.

\section{Neural network training}
\label{sec:learning}

In this section, we present the training of a neural network on a particular graph. 
The aim is for the neural network to learn how to approximate the behavior of a model-based planner  on the graph. To this end, we first use our solver to compute solutions for several random instances. Then we train the neural network for solving the same instances and use the previously generated solutions as supervision. We want to leverage the powerful mechanism of neural networks to guide our path planner. 
We first provide a brief introduction to neural networks and their main concepts.

\subsection{Neural Networks}
In recent years, neural networks (NNs), in particular deep neural networks, have achieved major breakthroughs in various areas of computer vision (image classification \cite{krizhevsky_2012}, \cite{simonyan_2014}, \cite{he_2016}, object detection \cite{ren_2015}, \cite{redmon_2016}, \cite{he_2017}, semantic segmentation \cite{long_2015}), neural machine translation \cite{sutskever_2014}, computer games \cite{silver_2016}, \cite{silver_2017} and many other fields. While the fundamental principles of training neural networks are known since many years, the recent improvements are due to a mix of availability of large image datasets, advances in GPU-based computation and increased shared community effort.

Deep neural networks enable multiple levels of abstraction of data by using models with millions of trainable parameters coupled with non-linear transformations of the input data. It is known that a sufficiently large neural network can approximate any continuous function \cite{funahashi_1989}, although the cost of training such a network can be prohibitive. With this in mind, we attempt to train a neural network to approximate the behavior of a model-based planner.

In spite of the complex structure of a NN, the main mechanism is straightforward. A \emph{feedforward neural network}, or \emph{multi-layer perceptron (MLP)}, with $L$ layers describes a function $f(\xb; \thetab): \real^{d_{x}} \mapsto \real^{d_{y}}$  that maps an input vector $\xb \in \real^{d_{x}}$ to an output vector $\yb \in \real^{d_{y}}$.  $\xb$ is the input data that we need to evaluate ($\eg$ an image, a signal, a graph, etc.), while $\yb$ is the expected decision from the NN ($\eg$ a class index, a heatmap, etc.). The function $f$ performs $L$ successive operations over the input $\xb$:
\begin{align}
  h^{(l)} = f^{(l)}(h^{(l-1)}; \theta^{(l)}), \qquad l=1,\dots,L
\label{eq:layers}
\end{align}
where $h^{(l)}$ is the hidden state of the network ($\ie$ features from intermediate layers) and  $f(h^{(l-1)}; \theta^{(l)}): \real^{d_{l-1}} \mapsto \real^{d_{l}}$ is the mapping function performed at layer $l$; $h_0=\xb$. In other words, $f(\xb)=f^{(L)}(f^{(L-1)}(\dots f^{(1)}(\xb)\dots))$. Each intermediate mapping depends on the output of the previous layer and on a set of trainable parameters $\theta^{(l)}$. We denote by ${\thetab=\{\theta^{(1)},\dots,\theta^{(L)}\}}$ the entire set of parameters of the network.
The intermediate functions $f(h^{(l-1)}; \theta^{(l)})$ have the form:
\begin{align}
f^{(l)}(h^{(l-1)}; \theta^{(l)}) = \sigma\left( \theta^{(l)} h^{(l-1)} + b^{(l)} \right) , 
\label{eq:linear}
\end{align}
where $\theta^{(l)}\in\real^{d_l\times d_{l-1}}$ and $b^{(l)}\in\real^{d_l}$ are the trainable parameters and the bias, while $\sigma(\cdot)$ is an \emph{activation} function which is applied individually to each element of its input vector and introduces non-linearities.
Intermediate layers are actually a combination of linear classifiers followed by a piecewise non-linearity from the activation function. Layers with this form are termed \emph{fully-connected layers}.

NNs are typically trained using labeled training data, $\ie$ a set of input-output  pairs $(\xb_i, \yb_i)$,  $i=1,\dots,N$, where $N$ is the size of the training set. During training we aim to minimize the training loss:
\begin{align}
\mathcal{L}(\thetab) = \frac1N\sum_{i=1}^N \ell(\hat{\yb}_i,\yb_i) ,
\label{eq:loss}
\end{align}
where $\hat{\yb_i}=f(\xb_i; \thetab)$ is the estimation of $\yb_i$ by the NN and ${\ell: \real^{d_L}\times \real^{d_L} \mapsto \real}$ is the loss function. $\ell$ measures the distance between the true label $\yb_i$ and the estimated one $\hat{\yb_i}$. Through \emph{backpropagation}, the information from the loss is transmitted to all $\thetab$ and gradients of each $\theta_l$ are computed w.r.t. the loss $\ell$. The optimal values of the parameters $\thetab$ are then found via stochastic gradient descent (SGD) which updates $\thetab$ iteratively towards the minimization of $\mathcal{L}$. The input data is randomly grouped into mini-batches and parameters are updated after each pass. The entire dataset is passed through the network multiple times and the parameters are updated after each pass until reaching a satisfactory optimum. 
In this manner all the parameters of the NN are learned jointly and the pipeline allows the network to learn to extract features and to learn other more abstract features on top of the lower layers.

Convolutional Neural Networks (CNNs) \cite{lecun_1995} are a generalization of multilayer perceptrons for 2D data. In convolutional layers, groups of parameters (which can be seen as small fully-connected layers) are slided across an input vector similarly to filters in image processing. This reduces significantly the number of parameters of the network since weights are \emph{shared} across locations, whereas in fully connected layers there is a parameter for each location in the input. Since the convolutional units act locally, the input to the network can have a variable size. A convolutional layer is also a combination of linear classifiers (\ref{eq:linear}) and the output of such layer is 2D and is called \emph{feature map}. CNNs are highly popular in most recent approaches for computer vision problems.

\subsection{Graph Convolutional Networks}
Graph Convolutional Networks (GCNs) are generalizations of CNNs to non-Euclidean graphs. GCNs are in fact neural networks based on local operators on a graph $\mathcal{G}=(\mathcal{V},\mathcal{E})$ which are derived from spectral graph theory. The filter parameters are typically shared over all locations in the graph, thus the name $convolutional$. 
In the past two years there has been a growing interest for transferring the intuitions and practices from deep neural networks on structured inputs to graphs \cite{gori_2005}, \cite{henaff_2015}, \cite{defferrard_2016}, \cite{kipf_2016}.
\cite{bruna_2013} and \cite{henaff_2015} bridge spectral graph theory with multi-layer neural networks by learning smooth spectral multipliers of the graph Laplacian. Then \cite{defferrard_2016} and \cite{kipf_2016} approximate these smooth filters in the spectral domain using polynomials of the graph Laplacian. The free parameters of the polynomials are learned by a neural network, avoiding the costly computation of the eigenvectors of the graph Laplacian.
We refer the reader to \cite{bronstein_2017} for a comprehensive review on deep learning on graphs.

We consider here the approach of \cite{kipf_2016}. The GCNs have the following layer propagation rule:
\begin{equation}
  \textstyle
  h^{(l+1)}= \sigma\!\left(\tilde{D}^{-\frac{1}{2}} \tilde{A}\tilde{D}^{-\frac{1}{2}}h^{(l)} \theta^{(l)} \right),
\label{eq:gcn-layer}
\end{equation}
where $\tilde{A} = A + I$ is the adjacency matrix of the graph with added self-connections such that when multiplying with $A$ we aggregate features vectors from both a node and its neighbors; $I$ the identity matrix. $\tilde{D}$ is the diagonal node degree matrix of $\tilde{A}$. $\sigma(\cdot)$ is the activation function, which we set to $\mathrm{ReLU}(\cdot) = \max(0,\cdot)$. 
The diagonal degree $D$ is employed for normalization of $A$ in order to avoid change of scales in the feature vectors when multiplying with $A$. \cite{kipf_2016} argue that using a symmetric normalization, \ie \  $\tilde{D}^{-\frac{1}{2}} \tilde{A}\tilde{D}^{-\frac{1}{2}}$, ensures better dynamics compared to simple averaging of neighboring nodes in one-sided normalization $\tilde{D}^{-1}A$.

The input of our model is a vector $\xb$, where $h^{(0)}=\xb$, containing information about the problem instance, including the graph representation. We describe next the chosen encoding for vector $\xb$.

\subsection{Vector encoding of instances}
For every instance $I = (s,d,M)$ of a given graph $\mathcal{G}=(\mathcal{V},\mathcal{E})$, we associate a vector $\xb$ made up of triplets features per node in $\mathcal{G}$, making up for a total of $3\left\vert{\mathcal{V}}\right\vert$ features. The three features for a node $j \in \mathcal{V}$ are:

\begin{itemize}

\item A start node feature \[
    s_{j} = 
\begin{cases}
    \ 1, & \text{if node $j$ is a start node in instance $I$} \\
    \ 0,   & \text{otherwise}
\end{cases}
\]
\item An end node feature \[
    e_{j} = 
\begin{cases}
    \ 1, & \text{if node $j$ is the end node in instance $I$} \\
    \ 0,   & \text{otherwise}
\end{cases}
\]
\item A mandatory node feature \[
    m_{j} = 
\begin{cases}
    \ 1, & \text{if node $j$ is a mandatory node in instance $I$} \\
    \ 0,   & \text{otherwise}
\end{cases}
\]
\end{itemize}

\noindent
Vector $\xb$ is the concatenation of these features:
\begin{equation}
\xb= (s_{1},e_{1},m_{1},s_{2},e_{2},m_{2}, ..., s_{\left\vert{\mathcal{V}}\right\vert},e_{\left\vert{\mathcal{V}}\right\vert},m_{\left\vert{\mathcal{V}}\right\vert})
\label{equ:embedding}
\end{equation}

\begin{figure}
\centering
\includegraphics[scale=0.30]{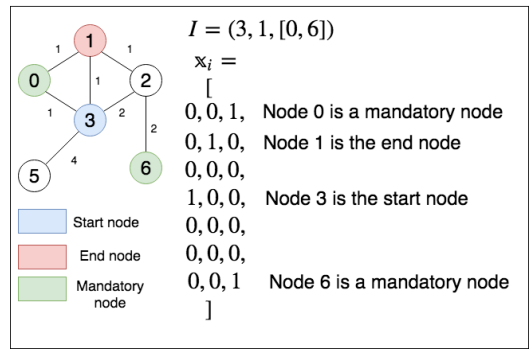}
\caption{Graph with seven nodes. The instance $I$ defined in the figure requires finding an optimal path from node 3 to node 1 which passes by node 0 and 6 at least once. $\xb$ is the vector associated with instance $I$. The number written next to an edge is its cost.}
\label{fig_vector}
\end{figure}

Figure \ref{fig_vector} illustrates an example of an instance $I$ in a graph and the associated vector $\xb$.

\subsection{Neural network architecture}

We define a neural network $f$ that uses a sequence of graph convolutions followed by a fully connected layer. The idea is to have the neural network take as input any instance $I$, and output a probability $\hat{\yb}$ over which node should be visited first from the start node in an optimal path that solves $I$. The weights of the neural network $\thetab$ are tuned in the training phase for this purpose. We train the neural network on instances that have already been solved with our solver. Here, the solver serves as a \emph{teacher} to the neural network and the neural network learns to approximate the results produced by the solver.

The output $\hat{\yb}$ of the neural network is a vector of size $\left\vert{\mathcal{V}}\right\vert$. We format the input vector $\xb$ as following:
we reshape $\xb$ into a matrix of size $(\left\vert{\mathcal{V}}\right\vert,3)$, which contains the start, end and mandatory features of every node by rows.
This matrix becomes the input for $f$, which aims to extract and aggregate local information over layers starting from the input. The output of the convolutions, a matrix, is flattened into a vector by concatenating its rows. A fully connected layer then links the flattened vector to a vector $z$ of real-numbers of size $\left\vert{\mathcal{V}}\right\vert$. Finally we use a softmax function to obtain a probability distribution. Formally, the softmax function is given by:
\begin{equation}
\text{softmax}(\mathbf {z} )_{i}={\frac {e^{z_{i}}}{\sum _{k=1}^{\left\vert{\mathcal{V}}\right\vert}e^{z_{k}}}}
\label{equ:softmax}
\end{equation}
This formula ensures that:
$\forall i \in \{1,2,..., \left\vert{\mathcal{V}}\right\vert\},  \text{softmax}(\mathbf {z} )_{i} > 0$ and $\sum _{i=1}^{\left\vert{V}\right\vert}\text{softmax} (\mathbf {z} )_{i} = 1$.

Once the training of the neural network is concluded, we expect it to return the next node to visit for any instance $I$ in the same manner as the solver.

\begin{figure}
\centering
\includegraphics[scale=0.25]{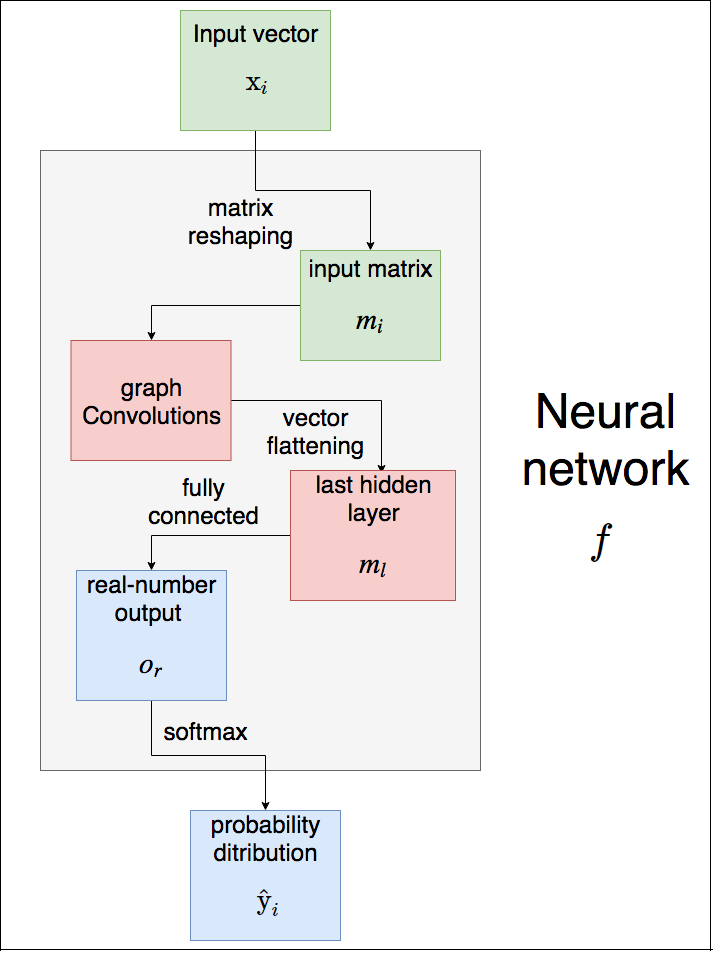}
\caption{Architecture of the neural network $f$. $f$ takes as input the vector $\xb_i$ of an instance $I_i$ and outputs a probability over all nodes in the graph, suggesting which node should be visited first.}
\label{fig_neural_network}
\end{figure}

\section{Data generation}
\label{sec:data_generation}

This section provides a global schema for data generation, either for the learning purpose, or for experimenting the global search strategy. 
We also give the main processing principles to conduct the learning phase.

\subsection{Number of instances}
Suppose the connected graph $\mathcal{G}=(\mathcal{V},\mathcal{E})$ has $\left\vert{V}\right\vert = n$ nodes. Let $P_n$ be the set of all existing instances for graph $\mathcal{G}$.
 The number of instances $C_n = \left\vert{P_n}\right\vert $ that exist is given by the following formula: \\ 
\begin{equation}
\label{number}
C_n = 2! {n\choose 2}\sum_{p=0}^{n-2} {n-2\choose p} \\
\end{equation}
where:

\begin{itemize}
\item $2!\tbinom {n}{2} = n(n-1)$ is the number of existing combinations of source-destination pairs, taking ordering into account.
\item $\sum_{p=0}^{n-2} {n-2\choose p}$ is, for every possible source-destination pair, the number of possible sets of mandatory nodes to visit. It can be simplified to $2^{n-2}$
\end{itemize}
Therefore, equation (\ref{number}) can be rewritten as:
\begin{equation}
C_n = 2^{n-2}n(n-1)\\
\end{equation}
The number of instances for a graph with $n=20$ is equal to $99\,614\,720$. It is reasonable to assume that calculating the solutions of all instances becomes too time-consuming.

\subsection{Instance generator}

In order to generate instances $I_i=(s_i,d_i,M_i)$, a generator function is built. 
It returns a set of instances $R$, all of which are solvable since the graph $\mathcal{G}$ is connected. 
These instances are sampled out of the set of all possible instance configurations $P_n=\{I_1, I_2, ..., I_{2^{n-2}n(n-1)}\}$ such that they cover $P_n$ as evenly as possible . 
For our experiments, we built 2 different connected graphs $\mathcal{G}_1$ and $\mathcal{G}_2$ containing respectively $15$ and $22$ nodes, based on the two benchmarks $b_1$ and $b_2$. Although these graphs are undirected, our work remains applicable to directed graphs.
However, the problem instances generated must remain realistic in terms of mission data. 
To be close to some 'realistic' instances, we first generate a shortest path length among all pairs of starting to ending nodes $(s_i, d_i)$ within the graph. 
We then apply a decimation ratio (typically 90\%), keeping the 10\% set of instances that have the longest paths.
For each resulting pair $(s_i, d_i)$, we then generate multiple random instances, with an increasing cardinality for the set of mandatory waypoints (typically from 1 to 10 mandatory waypoints). Each instance $I_i$ generated in $R$ is fed to our solver, providing an optimal $p_i$. The shortest path (Dijkstra) is set as probe, and for processing convenience, we set a 'time out' at 3 seconds for each instance. If no optimal solution can be proven, the instance is not considered for further learning process  (although some suboptimal solutions may be found).
Using this process, the number of instances generated and resolved are reported in the following table (\ref{tab:table0}):

\begin{table}[h!]
  \begin{center}
    \caption{Number of instances resolved under 3 seconds  by the solver out of the generation process, and used for the learning phase.}
    \label{tab:table0}
    \begin{tiny}
    \begin{tabular}{|l||r|r|r|r|r|r|r|}
      \hline  
      Mandatory waypoints \#: & 0 & 1&  2& 4& 6& 8& 10\\
      \hline
      \multicolumn{8}{|l|}{Instances for benchmark {\it maneuver} ($b_1$)  }\\
      generated (1256):& 42 & 212 & 246 & 252 & 252 & 252 & \\
      optimally solved (651):& 42 & 212 & 185 & 121 & 61 & 30 & \\
      \hline  
      \multicolumn{8}{|l|}{Instances for benchmark {\it exploration} ($b_2$) }\\
      generated (2503):& 69 & 368 & 410 & 414 & 414 & 414 & 414\\
      optimally solved (554):& 69 & 265 & 168 & 42 & 9 & 1 &  0\\ 
      \hline
    \end{tabular}
  \end{tiny}
 \end{center}
\end{table}

Only 651 (over 1256 generated) and 554 (over 2503) are solved optimally for respectively benchmarks $b_1$ and $b_2$. Indeed, easy instances are optimally solved more often than difficult ones. Given the number of nodes in the graphs, and applying (\ref{number}), we are working with less than 0.001\% of the total instances. As a consequence, most of the learning is performed on a small number of easy instances. For each instance $I_i=(s_i,d_i,M_i)$, let $\left\vert{M_i}\right\vert$ be the number of nodes in $M_i$.  The corresponding optimal path $p_i$ found by our solver is a path that passes through all the nodes in 

$M_i = \{m_{i1}, m_{i2}, ..., m_{i\left\vert{M_i}\right\vert}\}$, ie: $p_i = \{s_i, v_{i1}, v_{i2}, ..., v_{iq}, d_i\}$, $m_{ij} \in \{v_{i1}, v_{i2}, ..., v_{iq}\} | \forall j \in 1,2,...,\left\vert{M_i}\right\vert$

Since our neural network model predicts the next node to visit for a given instance $I$, our data is reprocessed before the training phase.  We use the following lemma to reprocess the data:

\newtheorem{ldef}{Lemma}

\begin{ldef}
\label{lemmaPath}
Let $I=(s,d,M)$ be a problem instance, and $p = \{s, v_1, v_2, v_3, ..., v_q, d\}$ an optimal path that solves $I$. It comes that:
\begin{itemize}
\item $\{v_1, v_2, v_3,..., v_q, d\}$ is an optimal solution for the instance $(v_1,d,M\backslash\{v_1\})$
\item $\{v_2, v_3, ..., v_q, d\}$ is an optimal solution for the instance $(v_2,d,M\backslash\{v_1, v_2\})$
\item ...
\item $\{v_q, d\}$ is an optimal solution for the instance $(v_q,d,M\backslash\{v_1, v_2, .., v_q\})$
\end{itemize}

More specifically: $\{v_i, v_{i+1}, ..., v_q, d\}$ is an optimal solution for the instance $(v_i,d,M\backslash\{v_1, v_2, ..., v_i\})| \forall i \in 1, 2, ..., q $

\end{ldef}

\begin{proof}
Suppose that $\{v_1, v_2, v_3, ..., v_q, d\}$ is not an optimal solution for $(v_1,d,M\backslash\{v_1\})$. There is therefore a path $p'$ that starts from $v_1$, ends in $d$, that visits every node in $M\backslash\{v_1\}$ with a lower cost than the path $\{v_1, v_2, v_3, ..., v_q, d\}$.\\
Therefore $p = \{s, v_1, v_2, v_3, ..., v_q, d\}$ is not optimal for the original problem $p$ since it does not take that shorter path, which is contradictory. It results that $\{v_1, v_2, v_3, ..., v_q, d\}$ is optimal for $(v_1,d,M\backslash\{v_1\})$. The same reasoning is applied recursively.
\end{proof}

\subsection{Data processing}

Let $(I_i,p_i)$ be an instance-solution pair that was previously generated, such that $I_i = (s_i, d_i, M_i)$ and $p_i = \{s_i, v_{i1}, v_{i2}, v_{i3}, ..., v_{iq_i}, d_i\}$.
This pair, which we call root pair, is split into several pairs $(I_{i,j},p_{i,j}) , \forall j \in 1,2,...,q_i$ in the same way as in Lemma \ref{lemmaPath}. This guarantees that for each instance $I_{i,j}$, $p_{i,j}$ is an optimal solution path. For each newly obtained pair ($I_{i,j}$, $p_{i,j}$), we store $(I_{i,j}, t_{i,j})$ in a dataset $d$, where $t_{i,j}\in V$ is the first node visited in path $p_{i,j}$ after the start node. The same process is applied for every root pair $(I_i,p_i)$ which was stored. 
The dataset is shuffled so as to compensate for the correlation resulting from splitting root pairs $(I_i,p_i)$ into pairs $(I_{i,j},p_{i,j})$, which are children instances and whose solutions enable the solving of the parent instance.

\subsection{Supervised learning}

Following the creation of dataset $d$, we train the neural network so that it can learn to approximate the behavior of our solver, and correctly predict the next step that should be taken in an optimal path for a given instance $I$. We take 80\% of the data in dataset $d$ for our training set. 
The test set is given by the remaining 20\% of the data in $d$.
Let $f$ be the function for the neural network defined previously. $f$ takes as input a vector instance $\xb_i$ and outputs a distribution vector $\hat{\yb}_i$ which is the probability distribution over all nodes in the graph of the next node to visit: $f(\xb_i; \thetab) = \hat{\yb}_i$ where $\thetab$ are the weights of the neural network.
We define the loss function $\mathcal{L}$ as the average of the logarithmic loss of the probabilities predicted by the neural network for each problem instance $I_i$ with the actual label target node $t_i$. 
$\mathcal{L}$ is defined as follows:

\begin{equation}
\mathcal{L}(\theta) = \frac{1}{m}\sum_{i=1}^{m}\sum_{j=1}^{n} -t_{ij}log(f(\xb_i;\thetab)_j)
\end{equation}

\begin{itemize}
\item $m$ is the number of training examples in the training set,
\item $n$ is the number of nodes in the graph,
\item $\xb_i$ is the vector of the problem instance $I_i$,
\item $t_{ij}$ is the variable that indicates whether for the problem instance $I_i$, the node $j$ was the next optimal node to visit: $t_{ij} = 1$ if so, else $t_{ij} = 0$,
\item $f(x_i;\thetab)_j$ is the probability the neural network outputs to visit node $j$ for the problem instance $I_i$.
\end{itemize}

The neural network is trained using a variant of the stochastic gradient descent (SGD) called \textit{Adam} \cite{diederik-kingma:adam}. \textit{Adam} uses adaptive learning rates for each variable of the neural network to decrease the number of steps required to minimize the loss function. To prevent overfitting, we apply the process of \textit{early stopping}: the training is stopped once the average loss on the test set stops decreasing, and restarts increasing. The training process is run in less than an hour for each graph on an nvidia DGX-station after parallelizing the gradient computation on 4 Tesla V100 GPUs. The processor (Intel Xeon E5-2698 v4 2.2 GHz) and the system memory (256 GB RDIMM DDR4) also contributed to fast data loading. We ran the training on the data obtained from graphs $\mathcal{G}_1$ and $\mathcal{G}_2$. Training curves are available in the appendix. 
The neural networks trained for each graph generalize relatively well on unknown instances (the neural network for graph $\mathcal{G}_1$ achieves 96\% accuracy its test set, while the neural network for graph $\mathcal{G}_2$ achieves 92\% accuracy on its test set).

\subsection{Framework and hyperparameters}

The neural network was implemented in python using Tensorflow. 
To build the tensorflow graph, the adjacency matrix is first loaded in the python environment, and is then used for the graph convolutions. 
In the training phase, we used a dropout layer which connects the last hidden layer to the output layer to further reduce overfitting. 
The probability of a neuron dropping out was selected to be $0.1$.
We used batch normalization \cite{ioffe-szegedy} for every layer in the graph convolutions with a decay parameter $\varepsilon = 0.9$. 
Lastly, we picked a learning rate $r=10^{-4}$ and trained the neural network with mini-batches of data of size $32$.

\section{Results and perspectives}
\label{sec:results}

\begin{figure*}
\centering
\includegraphics[scale=0.4]{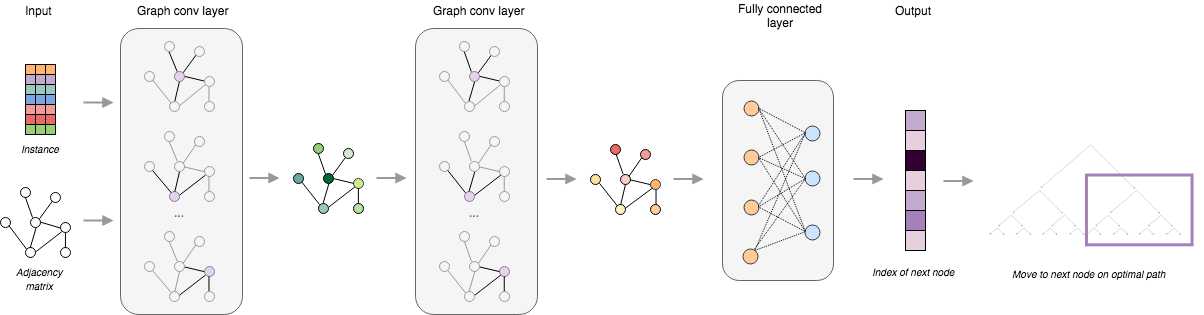}
\caption{Processing pipeline for path planning using GCNs. The GCN takes as input the adjacency matrix with costs and the instance. Graph convolutional layers process each node in the graph and its neighbors. In the hidden layers, new features are generated for each node in the graph. In the last layer, the features are passed through a fully-connected layer and a softmax. The softmax layer indicates the next node in the optimal path.}
\label{fig_gcn}
\end{figure*}

In this section, we evaluate to which extent the neural network can help solve new instances. When solving an instance $I$, the neural network is used at search start, and given as input the corresponding vector  $\xb$. The output $\hat{\yb}$ is used to create a preference ordering over all existing nodes in the graph, and corresponds to the preference of the next node to visit from the start node of the instance $I$. This order suggested by the neural network is provided to the solver at the root of the search tree, and children nodes of the root node will be visited in the order they appear in the preference ordering. Figure \ref{fig_gcn} depicts this process. The neural network is not used again to create preference orderings of root nodes of the resulting subtrees. We motivate this design choice by the fact that a large amount of operations is required for a feedforward pass. Probing with the neural network at every choice point would make the solving too slow and impractical.

We generate instances for two benchmarks, a maneuver benchmark, associated with graph $\mathcal{G}_1$, and an exploration benchmark, associated with graph $\mathcal{G}_2$. The first benchmark comprises 1008 instances, the second one 2208 instances. Note that those instances are generated with our instance generator, and therefore may contain anywhere from 0 mandatory nodes up to 10 mandatory nodes. We solve those instances using our original model-based planning solver without neural network support to obtain {\it reference} performance. Then, we evaluate solving performance on the same instances using a modified version of the solver based on neural network probing, that was trained over data provided in \S~\ref{sec:data_generation}. In both cases, we only keep results where the proof of optimality could be achieved. Results are reported in table \ref{tab:table01}, showing stable  improvements on all datasets of instances.

\begin{table}[h!]
  \begin{center}
    \caption{Number of instances resolved with proof of optimality. Comparison  between the reference version and the neural network probing one (under 3 seconds 'time out').}
    \label{tab:table01}
    \begin{small}
    \begin{tabular}{|l||r|r|r|r|}
      \hline  
      Mandatory waypoints \#: & 3 & 5&  7& 9\\
      \hline
      \multicolumn{5}{|l|}{Solving instances for benchmark {\it maneuver} ($b_1$)  }\\
      reference:& 167 & 99 & 43 & 15  \\
      neural net probing:& 220 & 185 & 131 & 89  \\
      \hline  
      \multicolumn{5}{|l|}{Solving instances for benchmark {\it exploration} ($b_2$) }\\
      reference:& 104 & 29 & 4 & 1 \\
      neural net probing:& 210 & 65 & 18 & 4 \\ 
      \hline
    \end{tabular}
  \end{small}
 \end{center}
\end{table}

Table \ref{tab:table1} summarizes the number of instances solved by both solvers under 3 seconds with more detailed search features. Average number of backtracks and solving time on {\it maneuver} are significantly lower, denoting an efficient pruning of the search tree. The situation is different for {\it exploration} with 10\% more backtracks. Given that the graph for {\it exploration} contains more nodes than the graph for {\it maneuver}, this result is explained by more complex instances solved optimally with neural network probing, with an accordingly high number of backtracks, while simple probing was unable to solve those instances, thus not taking into account the number of backtracks.

\begin{table}[h!]
  \begin{center}
    \caption{Global search features for proof of optimality, comparison between the reference version and the neural network probing one (under 3 seconds 'time out').}
    \label{tab:table1}
    \begin{small}
    \begin{tabular}{|l||r|r|}
      \hline  
      \textbf{Benchmarks} & {\it Maneuver} & {\it Exploration} \\
      \hline
      \multicolumn{3}{|l|}{Number of instances resolved}\\
      reference & 324 & 138\\
      neural net probing  & 625 & 297\\
      \hline
      \multicolumn{3}{|l|}{Average solving time for optimality}\\
      reference & 504 & 1044\\
      neural net probing  & 394 & 1203\\
      \hline
      \multicolumn{3}{|l|}{Average number of backtracks for optimality}\\
      reference & 55787 & 26065\\
      neural net probing  & 35176 & 26504\\
      \hline
    \end{tabular}
  \end{small}
 \end{center}
\end{table}

Higher performance could be obtained by training on more random instances. This would however require solving more instances with the initial solver before the training of the neural network, which is already time-consuming.

\section{Conclusion}
\label{sec:conclusion}

In order to solve AUGV path planning problems with mandatory waypoints, we introduced an algorithm capable of significantly accelerating the performance of a constraint-based solving method. The algorithm requires the solving of multiple random instances to enable the training of the neural network, which is then used to accelerate the solver itself on new instances. The approach is efficient, even with very small learning datasets, as required by application preparation requirements. The performance obtained is realistic on two representative AUGV planning benchmarks. A drawback is that the neural network can only be used to accelerate the solver on the same geometric graph from which the random instances were solved, making it impractical if a solution is required right away on a previously unknown graph. Nevertheless, if the graph is known in advance, it does become a far better alternative to trying to pre-compute the solution to all existing instances, the number of which grows exponentially with the number of nodes in the graph.

\section{ Acknowledgments}
We would like to thank Nicolas Fouquet for lending us a workstation for the duration of our experiments.

\nocite{kipf-welling:semi}
\nocite{diederik-kingma:adam}
\nocite{ioffe-szegedy}

\bibliographystyle{named}

\begin{small}
\bibliography{cnsp}
\end{small}

\clearpage

\appendix
\section{Appendix: Learning curves}
\subsection{Graph $\mathcal{G}_1$}
$\left\vert{V}\right\vert = 15$

\begin{figure}[!htb]
\begin{center}
\includegraphics[scale=0.28]{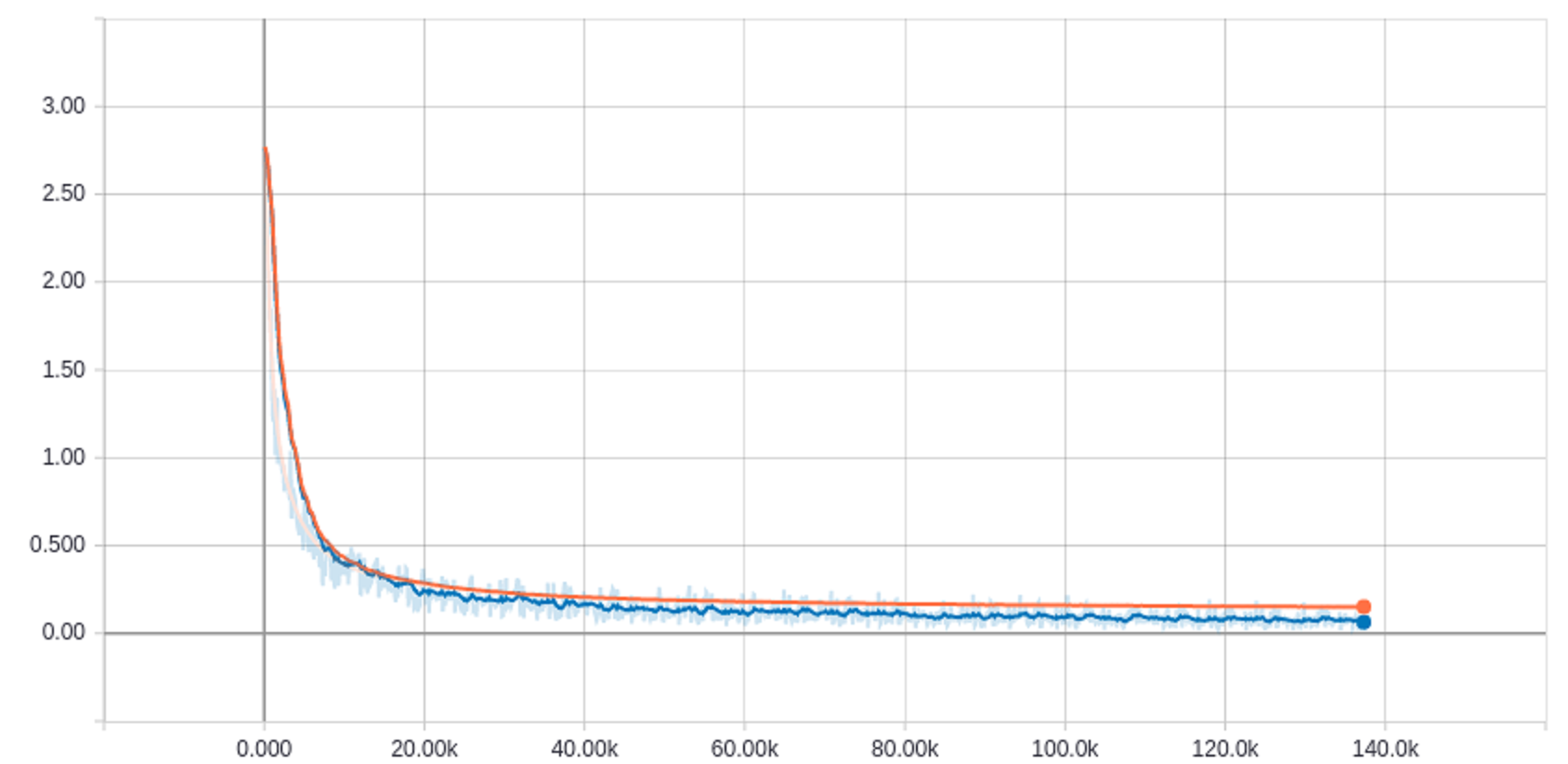}
\caption{{\it The loss function curve during training. The X-axis is the iteration step. The Y-axis is the average logarithmic log loss. The blue curve is the loss on a batch of training examples from the training set, the orange curve is the loss for the test set. The curves are smoothed by a  coefficient $\mu = 0.8$. We observe that the loss of the test set follows the loss of the training set as the training steps go on, while staying only slightly superior. This means the model, trained on instances of the training set, generalizes well on instances that are not part of the training set.}}
\label{crt_loss_curve}
\end{center}
\end{figure}

\begin{figure}[!htb]
\begin{center}
\includegraphics[scale=0.28]{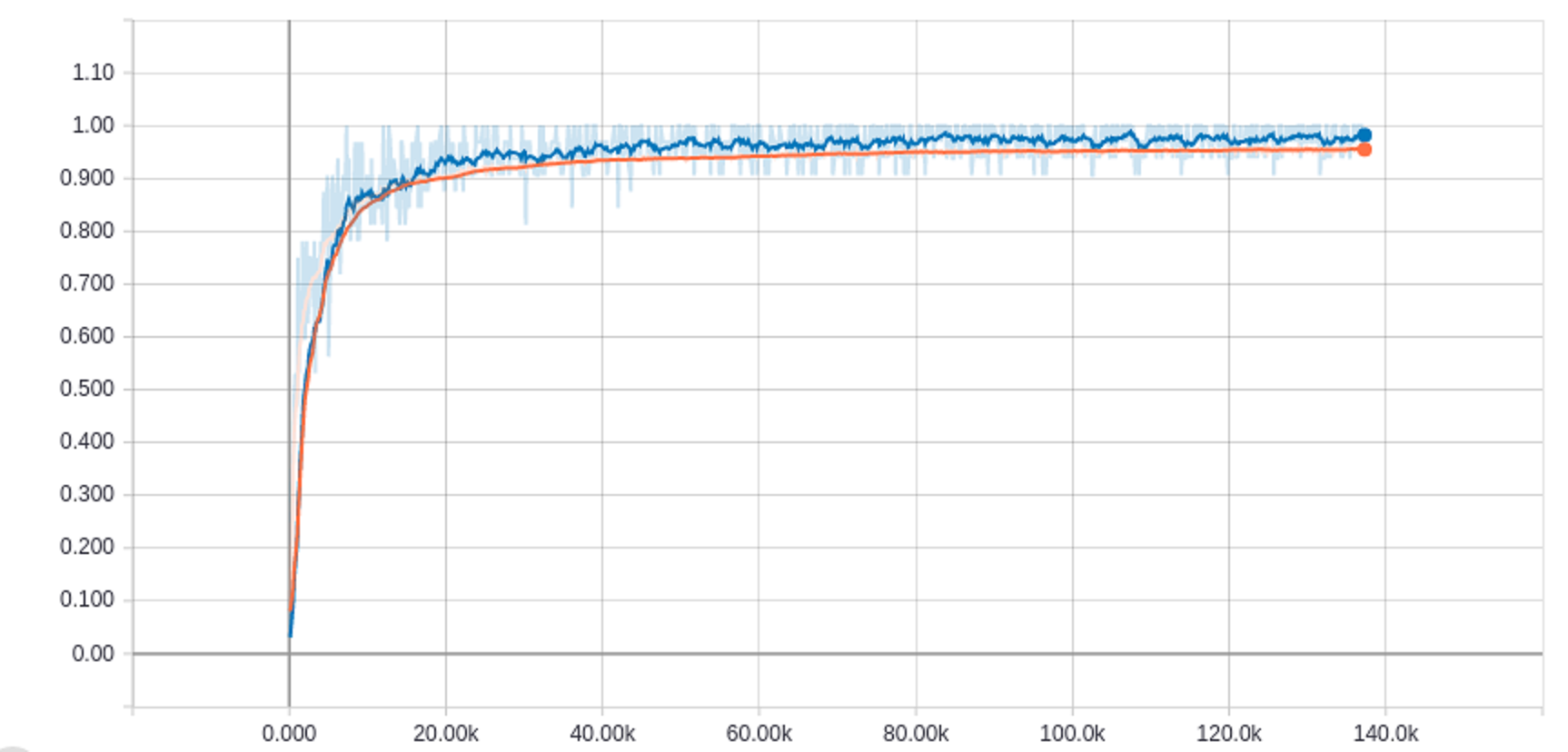}
\caption{{\it The accuracy curve during training. The X-axis is the iteration step. The Y-axis is the prediction accuracy. The blue curve is the accuracy on a batch of training examples from the training set, the orange curve is the accuracy for the test set. The curves are smoothed by a  coefficient $\mu = 0.8$. We observe that the accuracy of the test set follows the accuracy of the training set as the training steps go on, while staying only slightly inferior. This means the model, trained on instances of the training set, correctly predicts instances that are not part of the training set. By the end of the training phase, the model achieves an accuracy of 96\% on the test set.}}
\label{crt_accuracy_curve}
\end{center}
\end{figure}

\clearpage

\subsection{Graph $\mathcal{G}_2$}
$\left\vert{V}\right\vert = 22$

\begin{figure}[!htb]
\includegraphics[scale=0.28]{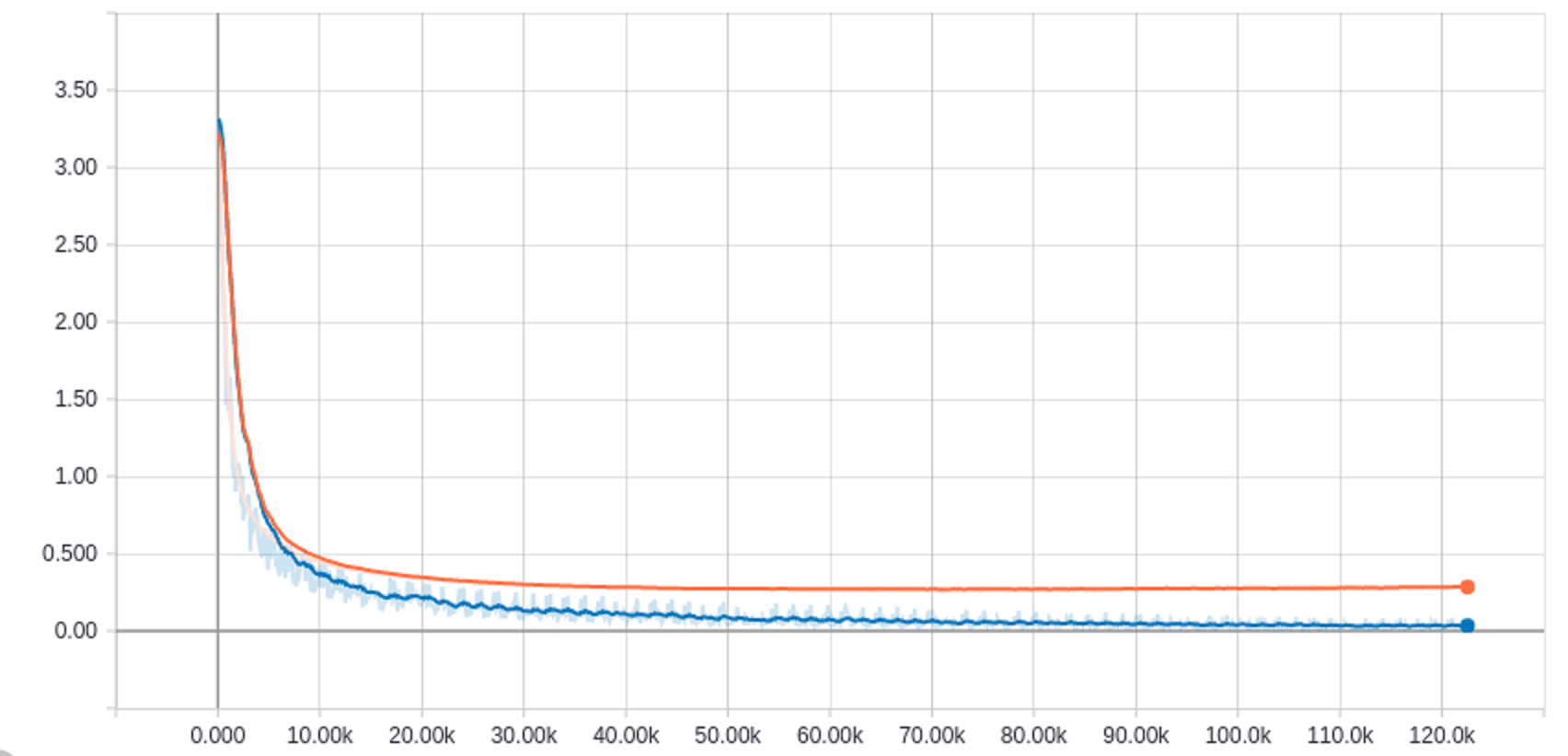}
\caption{{\it The loss function curve during training. The X-axis is the iteration step. The Y-axis is the average logarithmic log loss. The blue curve is the loss on a batch of training examples from the training set, the orange curve is the loss for the test set. The curves are smoothed by a  coefficient $\mu = 0.8$. We observe that the loss of the test set follows the loss of the training set as the training steps go on, while staying only slightly superior. This means the model, trained on instances of the training set, generalizes well on instances that are not part of the training set.}}
\label{troyes_loss_curve}
\end{figure}

\begin{figure}[!htb]
\begin{center}
\includegraphics[scale=0.28]{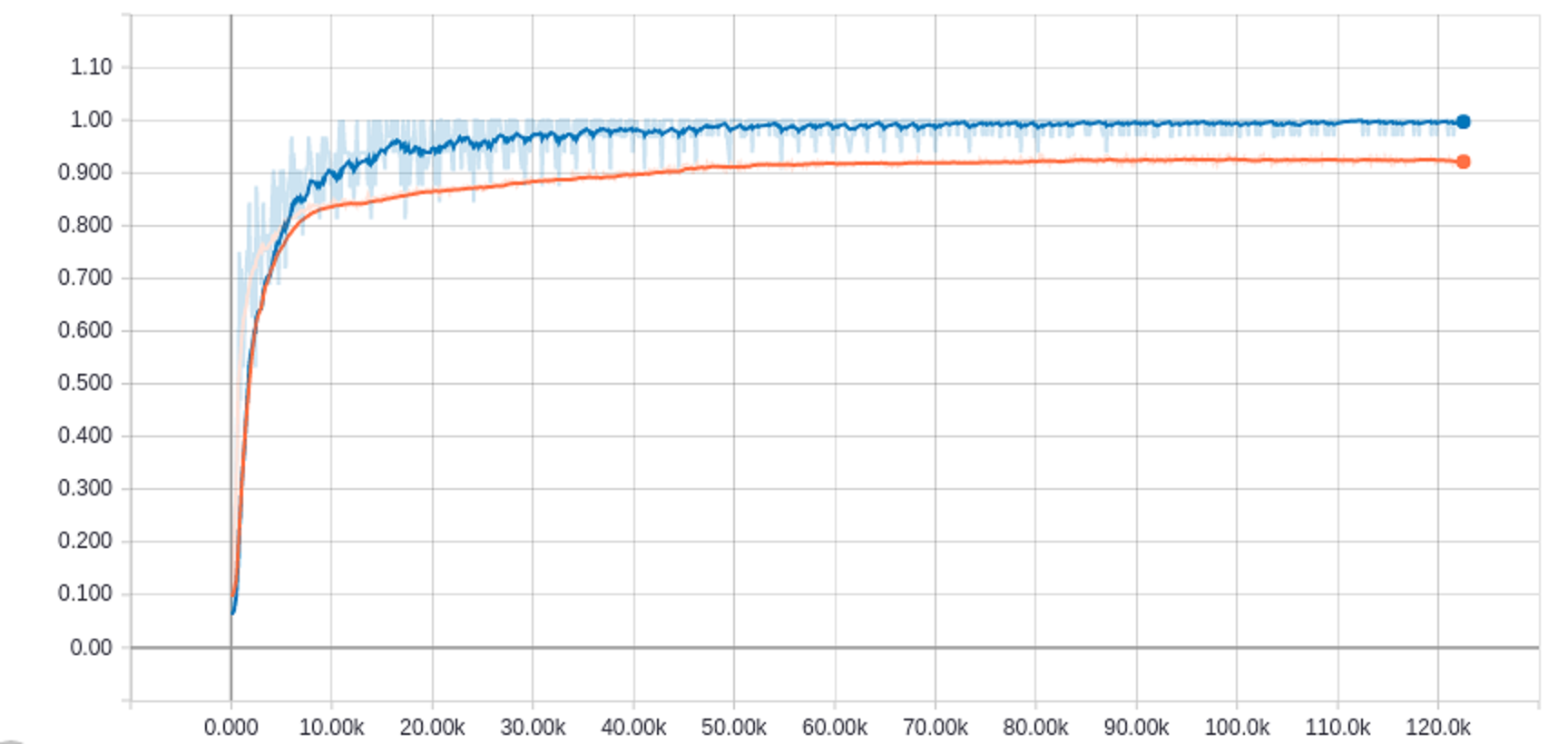}
\caption{{\it The accuracy curve during training. The X-axis is the iteration step. The Y-axis is the prediction accuracy. The blue curve is the accuracy on a batch of training examples from the training set, the orange curve is the accuracy for the test set. The curves are smoothed by a  coefficient $\mu = 0.8$. We observe that the accuracy of the test set follows the accuracy of the training set as the training steps go on, while staying only slightly inferior. This means the model, trained on instances of the training set, correctly predicts instances that are not part of the training set. By the end of the training phase, the model achieves an accuracy of 92\% on the test set.}}
\label{troyes_accuracy_curve}
\end{center}
\end{figure}

\end{document}